\documentclass{article}

\usepackage[preprint]{neurips_2019}

\usepackage{siunitx}
\usepackage[utf8]{inputenc} % allow utf-8 input
\usepackage[T1]{fontenc}    % use 8-bit T1 fonts
\usepackage{hyperref}       % hyperlinks
\usepackage{url}            % simple URL typesetting
\usepackage{booktabs}       % professional-quality tables
\usepackage{amsfonts}       % blackboard math symbols
\usepackage{nicefrac}       % compact symbols for 1/2, etc.
\usepackage{microtype}      % microtypography
\usepackage{color}
\usepackage{graphicx}
\usepackage{caption}
\usepackage{subcaption}
\usepackage{multirow}

\newcommand{\grover}[1]{{\it Selfie}}
\newcommand*\samethanks[1][\value{footnote}]{\footnotemark[#1]}

\title{\grover{}: Self-supervised Pretraining for \\ Image Embedding}
\author{Trieu H. Trinh\thanks{Equal contribution.} \hspace{2mm} Minh-Thang Luong\samethanks[1] \hspace{2mm} Quoc V. Le\samethanks[1] \\
  Google Brain \\
  \texttt{\{thtrieu,thangluong,qvl\}@google.com} \\
}

\begin{document}

\maketitle

\begin{abstract}
% We introduce \grover{}, a self-supervised pretraining technique that generalizes the concept of masked language modeling to continuous data, such as images.
We introduce a pretraining technique called \grover{}, which stands for 
% Filling in Missing Patches.
SELF-supervised Image Embedding. 
\grover{} generalizes the concept of masked language modeling of BERT~\citep{devlin2018bert} to continuous data, such as images, by making use of the Contrastive Predictive Coding loss~\citep{oord2018representation}.
Given masked-out patches in an input image, our method learns to select the correct patch, among other ``distractor'' patches sampled from the same image,  to fill in the masked location.
This classification objective sidesteps the need for predicting exact pixel values of the target patches.
The pretraining architecture of \grover{}
includes a network of convolutional blocks to process patches followed by an attention pooling network to summarize the content of unmasked patches before predicting masked ones. During finetuning, we reuse the convolutional weights found by pretraining.
We evaluate \grover{} on three benchmarks (CIFAR-10, ImageNet $32\times32$, and ImageNet $224\times224$) with varying amounts of labeled data, from 5\% to 100\% of the training sets. 
Our pretraining method provides consistent improvements to ResNet-50 across all settings compared to the standard supervised training of the same network. Notably, on ImageNet $224\times 224$ with 60 examples per class (5\%), our method improves the mean accuracy of ResNet-50 from 35.6\% to 46.7\%, an improvement of $11.1$ points in absolute accuracy. 
Our pretraining method also  improves ResNet-50 training stability, especially on low data regime, by significantly lowering the standard deviation of test accuracies across different runs.
\end{abstract}

\section{Introduction}
\label{sec:intro}

A weakness of neural networks is that they often require a large amount of labeled data to perform well. Although self-supervised/unsupervised representation learning~\citep{hinton2006fast,bengio2007greedy,raina2007self,vincent2010stacked} was attempted to address this weakness, most practical neural network systems today are trained with supervised learning~\citep{hannun2014deep,he2016deep,wu2016google}. Making use of unlabeled data through unsupervised representation learning to improve data-efficiency of neural networks remains an open challenge for the field.

Recently, language model pretraining has been suggested as a method for unsupervised representation learning in NLP~\citep{dai2015semi,ramachandran2016unsupervised,peters2018deep,howard2018universal,devlin2018bert}. 
Most notably, \citet{devlin2018bert} made an observation that bidirectional representations from input sentences are better than left-to-right or right-to-left representations alone. Based on this observation, they proposed the concept of masked language modeling by masking out words in a context to learn representations for text, also known as BERT. This is crucially achieved by replacing the LSTM architecture with the Transformer feedforward architecture~\citep{vaswani2017attention}.
The feedforward nature of the architecture makes BERT more ready to be applied to images. Yet BERT still cannot be used for images because images are continuous objects unlike discrete words in sentences. We hypothesize that bridging this last gap is key to translating the success of language model pretraining to the image domain.

In this paper, we propose a pretraining method called 
\grover{}, which stands for 
% Filling in Missing Patches.
SELF-supervised Image Embedding. 
% \grover{}\footnote{\grover{} stands for Generalized Reconstruction Objective for Visual Embedding Representations.} that
\grover{} generalizes BERT to continuous spaces, such as images. In \grover{}, we propose to continue to use classification loss because it is less sensitive to small changes in the image (such as translation of an edge) compared to regression loss which is more sensitive to  small perturbations. Similar to BERT, we mask out a few patches in an image and try to reconstruct the original image. To enable the classification loss, we sample  ``distractor'' patches  from the same image, and ask the model to classify the right patch to fill in a target masked location. Our method therefore can be viewed as a combination of BERT~\citep{devlin2018bert} and Contrastive Predictive Coding loss~\citep{oord2018representation}, where the negative patches are sampled from the same image, similar to Deep InfoMax~\citep{hjelm2018learning}. %. \textcolor{red}{Talk about Deep InfoMax as well.}

Experiments show that \grover{} works well across many datasets, especially when the datasets have a small number of labeled examples. On CIFAR-10, ImagetNet $32\times32$, and ImageNet $224\times224$, we observe consistent accuracy gains  as we vary the amount of labeled data from 5\% to 100\% of the typical training sets. The gain tends to be bigger when the labeled set is smaller. For example, on ImageNet $224\times224$ with only 60 labeled examples per class, pretraining using our method improves the mean accuracy of ResNet-50 by 11.1\%, going from 35.6\% to 46.7\%. Additional analysis on ImageNet $224\times224$ provides evidence that the benefit of self-supervised pretraining significantly takes off when there is at least an order of magnitude (10X) more unlabeled data than labeled data.

In addition to improving the averaged accuracy, pretraining ResNet-50 on unlabeled data also stabilizes its training on the supervised task. We observe this by reporting the standard deviation of the final test accuracy across 5 different runs for all experiments. On CIFAR-10 with 400 examples per class, the standard deviation of the final accuracy reduces 3 times comparing to training with the original initialization method. Similarly, on ImageNet rescaled to $32\times32$, our pretraining process gives an 8X reduction on the test accuracy variability when training on 5\% of the full training set. 
%Lastly, we will also discuss challenges in finding good pretrained models and caveats in our methods.

\section{Method}
\label{sec:method}

\begin{figure}[h!]
\centering
\includegraphics[width=1.0\textwidth]{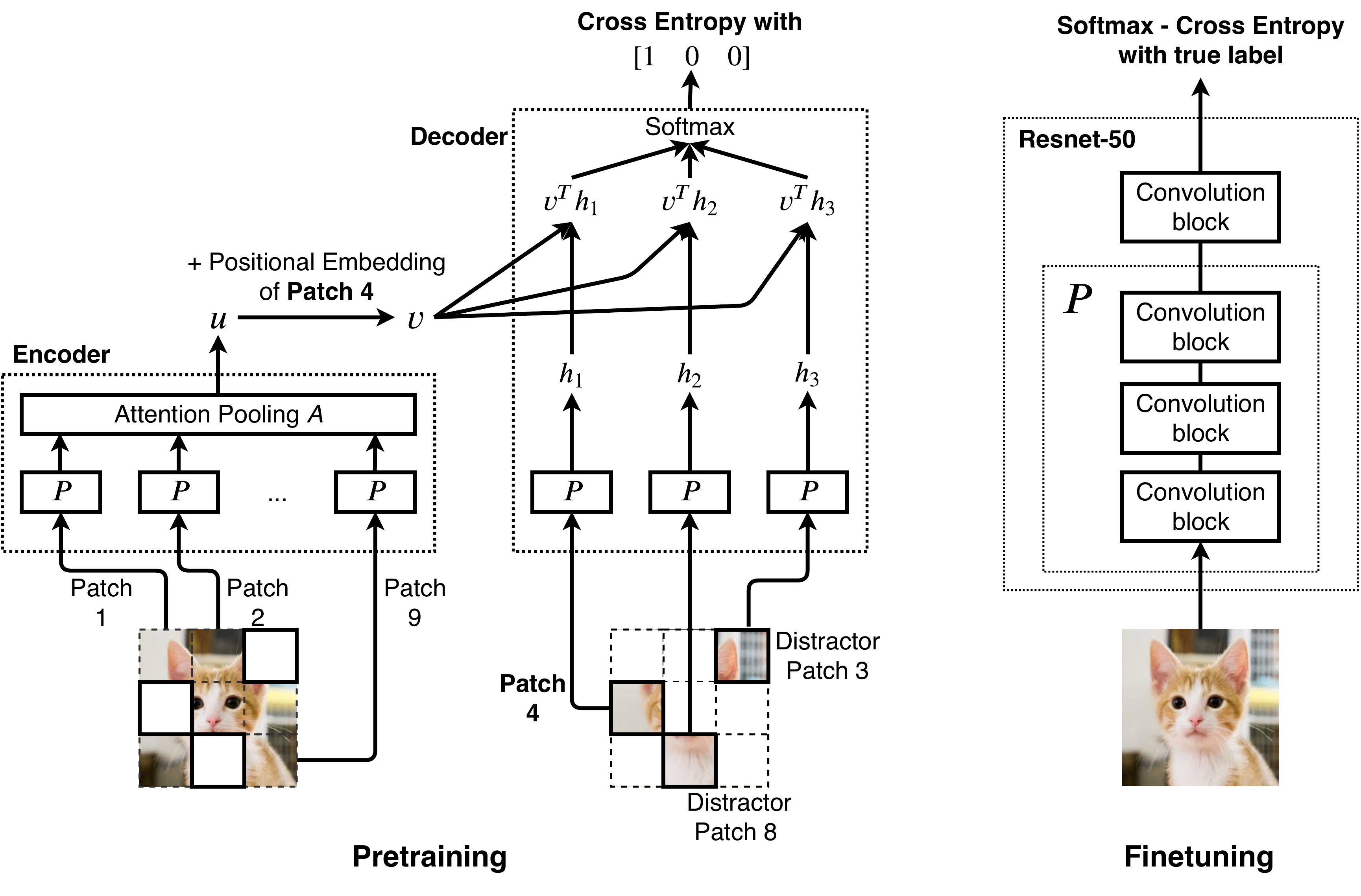}
\caption{
{\bf An overview of \grover{}}. 
{\it (Left) During pretraining}, our method makes use of an encoder-decoder architecture: the encoder takes in a set of square patches from the input image while the decoder takes in a different set. The {\it encoder} builds a single vector $u$ that represents all of its input patches using a patch processing network $P$
followed by an attention pooling network $A$.
The {\it decoder} then takes $u$ to predict its own input patches from their positions. Instead of predicting the actual pixel content, the decoder classifies the correct patch from negative examples (distractors) with a cross-entropy loss.
In our implementation, we use the first three blocks of ResNet-50 (equivalent to ResNet-36) for $P$ and Transformer layers~\citep{vaswani2017attention} for $A$.
Square patches are processed independently by $P$ to produce a feature vector per patch.
{\it (Right) During finetuning}, ResNet-50 is applied {\it on the full image}. Its first three blocks are initialized from the pretrained $P$, and the network is finetuned end-to-end.}
\label{fig:pretrain_loss}
\end{figure}

An overview of \grover{}
is shown in Figure~\ref{fig:pretrain_loss}. Similar to previous works in unsupervised/self-supervised representation learning, our method also has two stages: (1) Pretrain the model on unlabeled data and then (2) Finetune on the target supervised task. To make it easy to understand, let us first focus on the fine-tuning stage. In this paper, our goal is to improve ResNet-50, so we will pretrain the first three blocks of this architecture.\footnote{In our experiments, we found using the first three convolution blocks gives similar results to the full network (4 convolution blocks). During pretraining, therefore, only the first three blocks (i.e. ResNet-36) are used to save computation and memory load.} Let us call this network $P$. The pretraining stage is therefore created for training this $P$ network in an unsupervised fashion. 

Now let us focus on the pretraining stage. In the pretraining stage, $P$, a {\it patch processing} network, will be applied to small patches in an image to produce one feature vector per patch for both the encoder and the decoder.
In the {\it encoder}, the feature vectors are pooled together by an {\it attention pooling} network $A$ to produce a single vector $u$.
In the {\it decoder}, no pooling takes place; instead the feature vectors are sent directly to the computation loss to form an unsupervised classification task. The representations from the encoder and decoder networks are jointly trained during pretraining to predict what patch is being masked out at a particular location among other distracting patches.

In our implementation, to make sure the distracting patches are hard, we sample them from the same input image and also mask them out in the input image. Next we will describe in detail the interaction between the encoder and decoder networks during pretraining as well as different design choices.

\subsection{Pretraining Details} % implementation details}

\label{sec:pretrain}

The main idea is to use a part of the input image to predict the rest of the image during this phase. To do so, we first sample different square patches from the input. These patches are then routed into the encoder and decoder networks depending on whether they are randomized to be masked out or not. Let us take Figure~\ref{fig:pretrain_loss} as an example, where $Patch_1, Patch_2, Patch_5, Patch_6, Patch_7, Patch_9$ are sent into the encoder, whereas $Patch_3, Patch_4, Patch_8$ are sent into the decoder. 

All the patches are processed by the same patch processing network $P$.
On the {\it encoder} side, the output vectors produced by $P$ are routed into the attention pooling network to summarize these representations into a single vector $u$.  On the {\it decoder} side, $P$ creates output vectors $h_1$, $h_2$, $h_3$. The decoder then queries the encoder by adding to the output vector $u$ the location embedding of a patch, selected at random among the patches in the decoder (e.g., $location_4$) to create a vector $v$. The vector $v$ is then used in a dot product to compute the similarity between $v$ and each $h$. Having seen the dot products between $v$ and $h$'s, the decoder has to decide which patch is most relevant to fill in the chosen location (at $location_4$). %The cross entropy loss will be used to train both the Encoder and Decoder networks.
The cross entropy loss is applied for this classification task, whereas the encoder and decoder are trained jointly with gradients back-propagated from this loss.

During this pretraining process, the encoder network learns to compress the information in the input image to a vector $u$ such that when seeded by a location of a missing patch, it can recover that patch accurately. To perform this task successfully, the network needs to understand the global content of the full image, as well as the local content of each individual patch and their relative relationship. This ability proves to be useful in the downstream task of recognizing and classifying objects.

\paragraph{Patch sampling method.}  On small images of size $32\times 32$, we use a patch size of $8$, while on larger images of size $224\times224$, we use a patch size of $32\times32$. The patch size is intentionally selected to divide the image evenly, so that the image can be cut into a grid as illustrated in Figure~\ref{fig:pretrain_loss}. To add more randomness to the position of the image patches, we perform zero padding of 4 pixels on images with size $32\times32$ and then random crop the image to its original size.

\begin{figure}[ht!]
\centering
\includegraphics[width=0.9\textwidth]{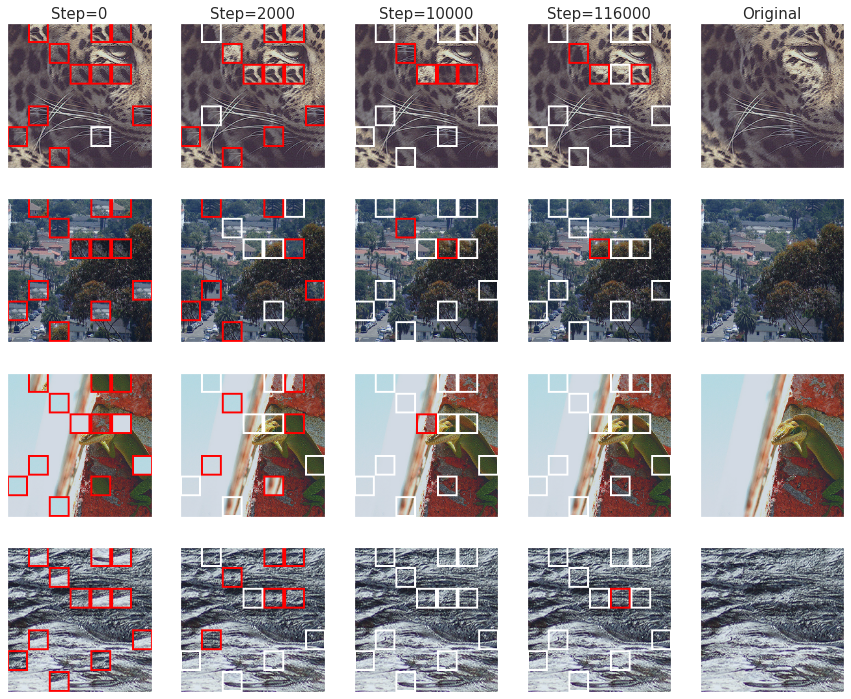}
\caption{From left to right: Improvement in predictions during our pretraining process on ImageNet $224\times224$. The patch size for this dataset is $32\times32$, which resulted in a grid of $7 \times 7$ patches. The masked-out patches are highlighted with a border of color white or red. The model is trained to put the masked-out patches back into their original slots. A border of color red indicates wrong prediction from the model. Here we display four different samples with the same masking positions fixed throughout the training process. At the beginning, the orders of the patches are mostly incorrect due to random initialization of the model. During training, the model learns to classify more correctly. As pretraining progresses from left to right, the model makes less error, while the mistakes made in later stages usually confuse between patches that have similar content. For example, the generic texture of the sky (row 3, step 2K), water (row 4, step 116K) or trees (row 2, step 10K) are generally interchangeable across locations.}
\label{fig:jigsaw}
\end{figure}

\paragraph{Patch processing network.} In this work, we focus on improving ResNet-50~\citep{he2016deep} on various benchmarks by pretraining it on unlabeled data. For this reason, we use ResNet-50 as the patch processing network $P$.\footnote{Our implementation of ResNet-50 achieves 76.9 $\pm$ 0.2 \% top-1 accuracy on ImageNet, which is in line with other results reported in the literature~\citep{he2016deep,zagoruyko2016wide,huang2017densely}.} 
As described before, only the first three blocks of ResNet-50 is used. Since the goal of $P$ is to reduce any image patch into a single feature vector, we therefore perform average pooling across the spatial dimensions of the output of ResNet-36.

\paragraph{Efficient implementation of mask prediction.} For a more efficient use of computation, the decoder is implemented to predict multiple correct patches for multiple locations at the same time. For example, in the example above, besides finding the right patch for $location_4$, the decoder also tries to find the right patch for $location_3$ as well as $location_8$. This way, we reuse three times as much computation from the encoder-decoder architecture. Our method is, therefore, analogous to solving a jigsaw puzzle where a few patches are knocked out from the image and are required to be put back to their original locations. This procedure is demonstrated in Figure~\ref{fig:jigsaw}.

\subsection{Attention Pooling}
\label{sec:attn}

In this section, we describe in detail the attention pooling network $A$ introduced in Section~\ref{sec:pretrain} and the way positional embeddings are built for images in our work.

\paragraph{Transformer as pooling operation.} We make use of Transformer layers to perform pooling. Given a set of input vectors $\{h_1, h_2, ..., h_n\}$ produced by applying the patch processing network $P$ on different patches, we want to pool them into a single vector $u$ to represent the entire image. There are multiple choices at this stage including max pooling or average pooling. Here, we consider these choices special cases of the attention operation (where the softmax has a temperature approaching zero or infinity respectively) and let the network learn to pool by itself. To do this, we learn a vector $u_0$ with the same dimension with $h$'s and feed them together through the Transformer layers:

$$u, h^{output}_1, h^{output}_2, ..., h^{output}_n = \textrm{TransformerLayers}(u_o, h_1, h_2, .., h_n)$$

The output $u$ corresponding to input $u_o$ is the pooling result. We discard $h^{output}_i \ \ \forall i$.

\paragraph{Attention block.} Each self-attention block follows the design in BERT~\citep{devlin2018bert} where self-attention layer is followed with two fully connected layers that sequentially project the input vector to an intermediate size and back to the original hidden size. The only non-linearity used is GeLU and is applied at the intermediate layer. We perform dropout with rate $0.1$ on the output, followed by a residual connection connecting from the block's input and finally layer normalization.

\paragraph{Positional embeddings.} For images of size $32\times32$, we learn a positional embedding vector for each of the 16 patches of size $8\times8$. Images of size $224\times224$, on the other hand, are divided into a grid of $7\times7$ patches of size $32\times32$. Since there are significantly more positions in this case, we decompose each positional embedding into two different components: row and column embeddings. The resulting embedding is the sum of these two components. For example, instead of learning 49 positional embeddings, we only need to learn $7 + 7 = 14$ positional embeddings. This greatly reduces the number of parameters and helps with regularizing the model.

\subsection{Finetuning Details} % implementation details}

As mentioned above, in this phase, the first three convolution blocks of ResNet-50 is initialized from the pretrained patch processing network. The last convolution block of ResNet-50 is initialized by the standard initialization method. ResNet-50 is then applied on {\it the full image} and finetuned end-to-end.

\section{Experiments and Results}
\label{sec:exp}

In the following sections, we investigate the performance of our proposed pretraining method, \grover{}, on standard image datasets, such as CIFAR-10 and ImageNet. To simulate the scenario when we have much more unlabeled data than labeled data, we sample small fractions of these datasets and use them as labeled datasets, while the whole dataset is used as unlabeled data for the pretraining task.

\subsection{Datasets}

We consider three different datasets: CIFAR-10, ImageNet resized to $32\times32$, and ImageNet original size ($224\times224$). For each of these datasets, we simulate a scenario where an additional amount of unlabeled data is available besides the labeled data used for the original supervised task. For that purpose, we create four different subsets of the supervised training data with approximately 5\%, 10\%, 20\%, and 100\% of the total number of training examples. 
On CIFAR-10, we replace the 10\% subset with one of 4000 training examples (8\%), as this setting is used in~\citep{oliver2018realistic,cubuk2018autoaugment}. In all cases, the whole training set is used for pretraining (50K images for CIFAR-10, and 1.2M images for ImageNet).

\subsection{Experimental setup}

\paragraph{Model architecture.} We reuse all settings for ResNet convolution blocks from ResNet-50v2 including hidden sizes and initialization~\citep{he2016identity}. Batch normalization is performed at the beginning of each residual block. For self-attention layers, we apply dropout on the attention weights and before each residual connection with a drop rate of 10\%. 
The sizes of all of our models are chosen such that each architecture has roughly 25M parameters and 50 layers, the same size and depth of a standard ResNet-50. For attention pooling, three attention blocks are added with a hidden size of $1024$, intermediate size $640$ and $32$ attention heads on top of the patch processing network $P$.

\paragraph{Model training.} Both pretraining and finetuning tasks are trained using Momentum Optimizer with Nesterov coefficient of $0.9$. We use a batch size of $512$ for CIFAR-10 and $1024$ for ImageNet. Learning rate is scheduled to decay in a cosine shape with a warm up phase of 100 steps and the maximum learning rate is tuned in the range of $[0.01, 0.02, 0.05, 0.1, 0.2, 0.4]$. We do not use any extra regularization besides an $L2$ weight decay of magnitude $0.0001$. The full training is done in $120,000$ steps.
Furthermore, as described in Section~\ref{sec:pretrain}, we divide the images into non-overlapping square patches of size $8 \times 8$ or $32\times 32$ during pretraining and sample a fraction $p$ of these patches to predict the remaining. We try for two values of $p$: 75\% or 50\% and tune it as a hyper-parameter.

\paragraph{Reporting results.} For each reported experiment, we first tune its hyper-parameters by using 10\% of training data as validation set and train the neural net on the remaining 90\%. Once we obtain the best hyper-parameter setting, the neural network is retrained on 100\% training data 5 times with different random seeds. We report the mean and standard deviation values of these five runs.

\subsection{Results}
\label{sec:results}
\begin{figure}[h!]
\centering
\includegraphics[width=1.0\textwidth]{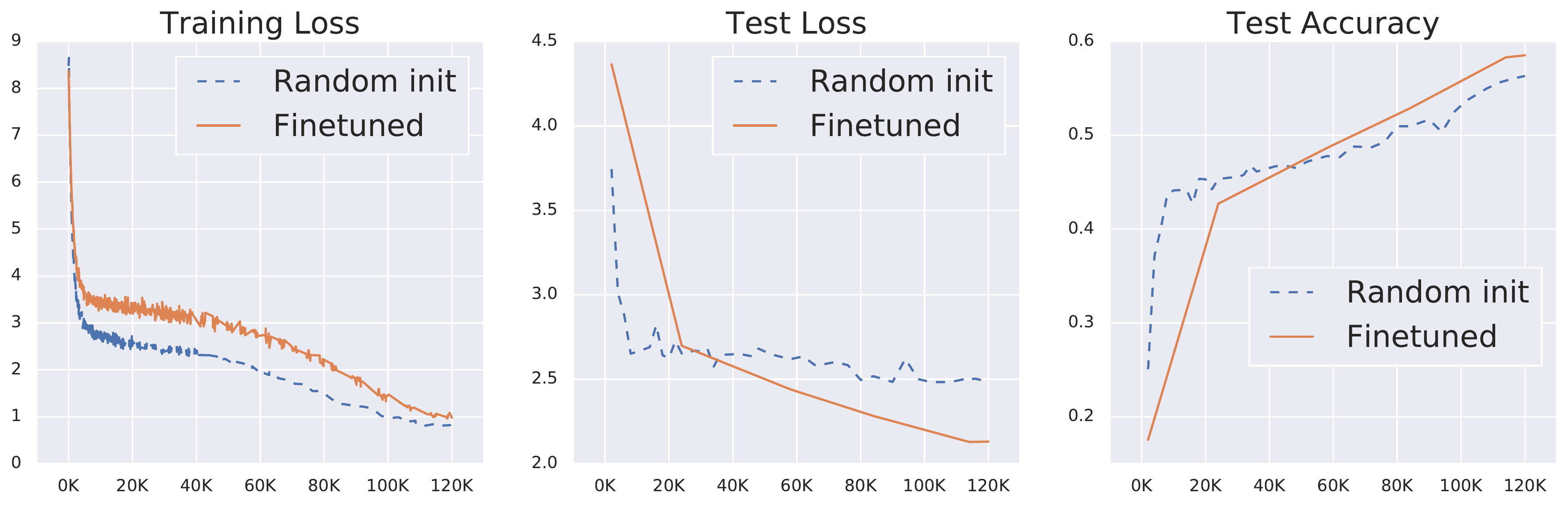}
\caption{Regularization effect of our pretraining method on the 10\% subset of Imagenet $224\times224$. We observe the values of training loss, test loss and test accuracy during 120K steps of supervised training, comparing a randomly initialized model to one that is initialized from pretrained weights.}
\label{fig:compare}
\end{figure}

We report the accuracies with and without pretraining across different labeled dataset sizes in Table~\ref{tab:result}.  
As can be seen from the table, \grover{} yields 
consistent improvements in test accuracy across all three benchmarks (CIFAR-10, ImageNet $32\times32$, ImageNet $224\times224$) with varying amounts of labeled data.
Notably, on ImageNet $224\times224$, a gain of 11.1\% in absolute accuracy is achieved when we use only 5\% of the labeled data. We find the pretrained models usually converge to a higher training loss, but generalizes significantly better than model with random initialization on test set. This highlights the strong effect of regularization of our proposed pretraining procedure. An example is shown in Figure~\ref{fig:compare} when training on 10\% subset of Imagenet$224\times224$.

Beside the gain in mean accuracy, training stability is also enhanced as evidenced by the reduction in standard deviation in almost all experiments. When the unlabeled dataset is the same with the labeled dataset (Labeled Data Percentage = 100\%), the gain becomes  small as expected. 

\begin{table}[!ht]
  \caption{Test accuracy (\%) of ResNet-50 with and without pretraining across datasets and sizes.}
  \label{tab:result}
  \centering\small
  \begin{tabular}{cccccc}
    \toprule
    & & \multicolumn{4}{c}{Labeled Data Percentage} \\
    \cline{3-6}
    & & {\it 5\%} & {\it 8\%} & {\it 20\%} & {\it 100\%} \\
    \cline{3-6}
    \multirow{3}{*}{CIFAR-10} & Supervised & 
    75.9 $\pm$ 0.7 & 79.3 $\pm$ 1.0  & 88.3 $\pm$ 0.3 & 95.5 $\pm$ 0.2 \\
    % & Pretrained & 
    & \grover{} Pretrained & 
    75.9 $\pm$ 0.4 & 80.3 $\pm$ 0.3 & 89.1 $\pm$ 0.5 & 95.7 $\pm$ 0.1 \\
    & $\Delta$ & 0.0 & \textbf{+1.0} & \textbf{+0.8} & \textbf{+0.2} \\ 
    \hline
    & & {\it 5\%} & {\it 10\%} & {\it 20\%} & {\it 100\%} \\
    \cline{3-6}
    \multirow{3}{*}{ImageNet $32\times32$} & Supervised & 
    13.1 $\pm$ 0.8 & 25.9 $\pm$ 0.5 & 32.7 $\pm$ 0.4 & 55.7 $\pm$ 0.6 \\
    % & Pretrained &
    & \grover{} Pretrained & 
    18.3 $\pm$ 0.1 & 30.2 $\pm$ 0.5 & 33.5 $\pm$ 0.2 & 56.4 $\pm$ 0.6 \\
    & $\Delta$ & \textbf{+5.2} & \textbf{+4.3} & \textbf{+0.8} & \textbf{+0.7} \\
    \hline
    \multirow{3}{*}{ImageNet $224\times224$} & Supervised & 
    35.6 $\pm$ 0.7 & 59.6 $\pm$ 0.2  & 65.7 $\pm$ 0.2 & 76.9 $\pm$ 0.2 \\
    % & Pretrained & 
    & \grover{} Pretrained & 
    46.7 $\pm$ 0.4 & 61.9 $\pm$ 0.2 & 67.1 $\pm$ 0.2 & 77.0 $\pm$ 0.1 \\
    & $\Delta$ & \textbf{+11.1} & \textbf{+2.3} & \textbf{+1.4} & \textbf{+0.1} \\ 
    \bottomrule
  \end{tabular}
\end{table}

\paragraph{Baseline Comparison.} We want to emphasize that our ResNet baselines are very strong compared to those in \citep{he2016deep}. Particularly, on CIFAR-10, our ResNet with pure supervised learning on 100\% labeled data achieves 95.5\% in accuracy, which is better than the accuracy 94.8\% achieved by DenseNet~\citep{huang2017densely} and close to 95.6\% obtained by Wide-ResNet~\citep{zagoruyko2016wide}.
Likewise, on ImageNet $224\times224$, our baseline reaches 76.9\% in accuracy, which is on par with the result reported in \citep{he2016deep}, and surpasses the 76.2\% accuracy of DenseNet~\citep{huang2017densely}. Our pretrained models further improve on our strong baselines.

\paragraph{Contrast to Other Works.} 
Notice that our classification accuracy of 77.0\% on ImageNet $224\times224$ is also significantly better than previously reported results in unsupervised representation learning~\citep{pathak2016context,oord2018representation,kolesnikov2019revisiting}. For example, in a comprehensive study by~\citep{kolesnikov2019revisiting}, the best accuracy on ImageNet of all pretraining methods is around 55.2\%, which is well below the accuracy of our models.  Similarly, the best accuracy reported by Context Autoencoders~\citep{pathak2016context} and Contrastive Predictive Coding~\citep{oord2018representation}  are 56.5\% and 48.7\% respectively. We suspect that such poor performance is perhaps due to the fact that past works did not finetune into the representations learned by unsupervised learning. 

Concurrent to our work, there are also other attempts at using unlabeled data in semi-supervised learning settings. \citet{cpc19} showed the effectiveness of pretraining in low-data regime using cross-entropy loss with negative samples similar to our loss. However, their results are not comparable to ours because they employed a much larger network, ResNet-171, compared to the ResNet-50 architecture that we use through out this work. Consistency training with label propagation has also 
achieved remarkable % much better 
results. For example, the recent Unsupervised Data Augmentation~\citep{xie19} reported 94.7\% accuracy on the 8\% subset of CIFAR-10. We expect that ur self-supervised pretraining method can be  combined with label propagation to provide additional gains, as shown in~\citep{zhai2019s}.

\paragraph{Finetuning on ResNet-36 + attention pooling.} In the previous experiments, we finetune ResNet-50, which is essentially ResNet-36 and one convolution block on top, dropping the attention pooling network used in pretraining. We also explore finetuning on ResNet-36 + attention pooling and find that it slightly outperforms finetuning on ResNet-50 in some cases.\footnote{We chose to use ResNet-50 for finetuning as it is faster and facilitates better comparison with past works.}
More in Section~\ref{sec:modela}.

\paragraph{Finetuning Sensitivity and Mismatch to Pretraining.}
Despite the encouraging results, we found that there are difficulties in transferring pretrained models across tasks such as from ImageNet to CIFAR. For the 100\% subset of Imagenet $224\times224$, additional tuning of the pretraining phase using a development set is needed to achieve the result reported in Table~\ref{tab:result}. There is also a slight mismatch between our pretraining and finetuning settings: during pretraining, we process image patches independently whereas for finetuning, the model sees an image as a whole. We hope to address these concerns in subsequent works.

\section{Analysis}

\subsection{Pretraining benefits more when there is less labeled data}

\begin{figure}[h!]
\centering
\includegraphics[width=0.9\textwidth]{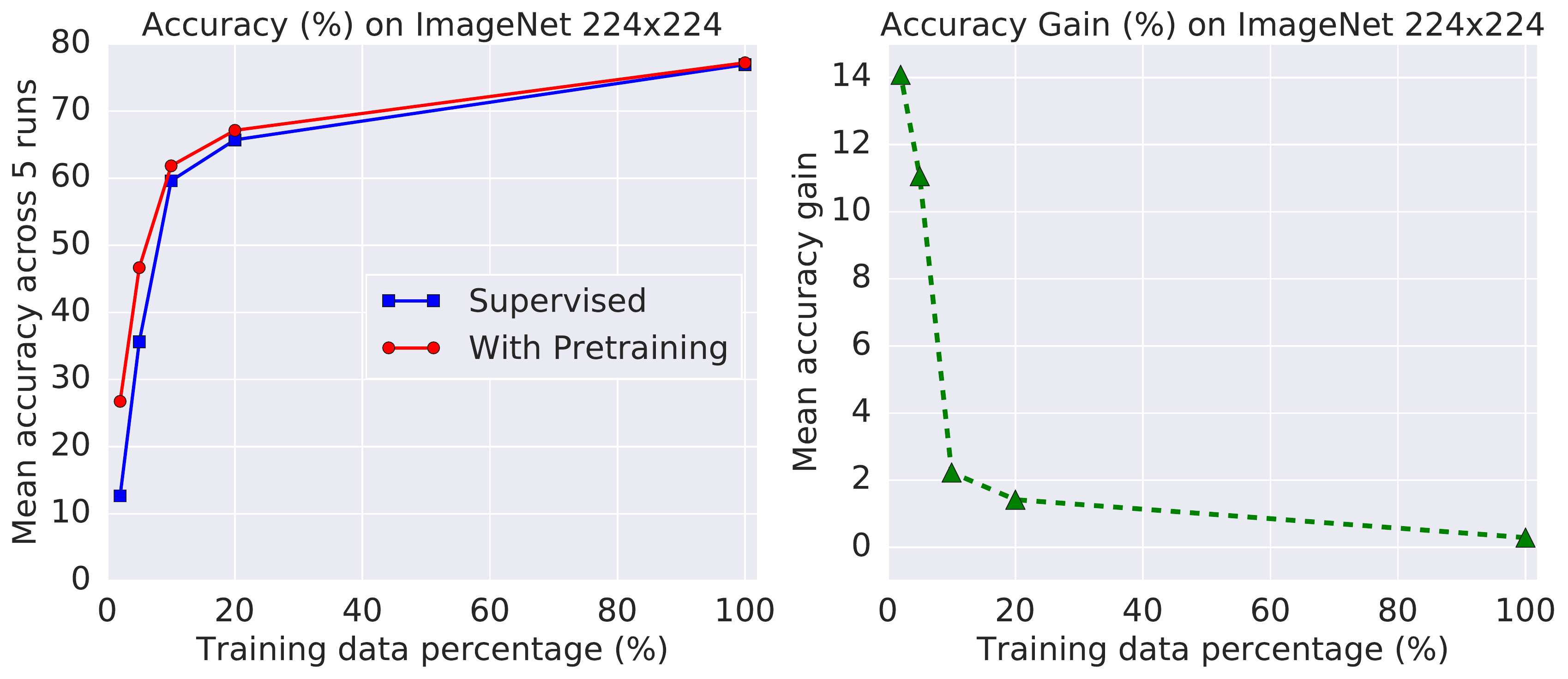}
\caption{
Pretraining with \grover{}
benefits the most when there is much more unlabeled data than labeled data. \textbf{Left:} Mean accuracy across five runs on ImageNet $224\times224$ for purely supervised model versus one with pretraining. \textbf{Right:} Mean accuracy gain from pretraining. The improvement quickly diminishes at the 10\% mark when there is 10 times more data than the labeled set.}
\label{fig:gain}
\end{figure}

In this section, we conduct further experiments to better understand our method, \grover{}, especially how it performs as we decrease the amount of labeled data. 
To do so, we evaluate test accuracy when finetuning on 2\%, 5\%, 10\%, 20\% and 100\% subset of ImageNet $224\times224$, as well as the accuracy with purely supervised training at each of the five marks. Similar to previous sections, we average results across five different runs for a more stable assessment. As shown in Figure~\ref{fig:gain}, the ResNet mean accuracy improves drastically when there is at least an order of magnitude more unlabeled image than the labeled set (i.e., finetuning on the 10\% subset). With less unlabeled data, the gain quickly diminishes. At the 20\% mark there is still a slight improvement of 1.4\% mean accuracy, while at the 100\% mark the positive gain becomes minimal, 0.1\%.

\subsection{Self-attention as the last layer helps finetuning performance.}
\label{sec:modela}

\begin{figure}[h!]
\centering
\includegraphics[width=0.605\textwidth]{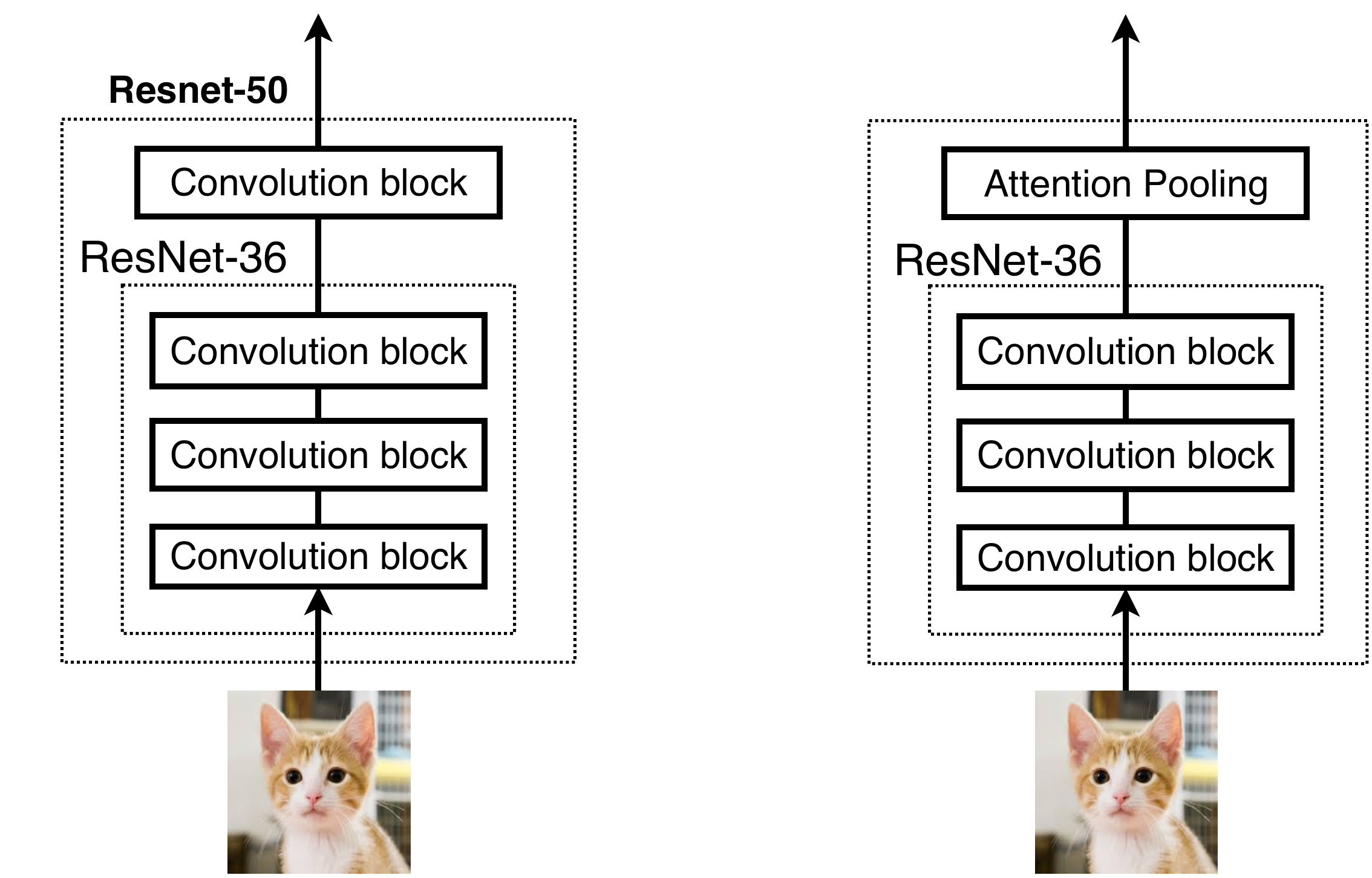}
\caption{{\it (Left)} ResNet-50 architecture. {\it (Right)} ResNet-36 + attention pooling architecture.}
\label{fig:resnet36}
\end{figure}

As mentioned in Section~\ref{sec:results}, we explore training ResNet-36 + attention pooling (both are reused from pretraining phase) on CIFAR-10 and ImageNet $224\times 224$ on two settings: limited labeled data and full access to the labeled set. The architectures of the two networks are shown in Figure~\ref{fig:resnet36}. Experimental results on these two architectures with and without pretraining are reported in Table~\ref{tab:modela}.

\begin{table}[h!]
\small
  \caption{Accuracy (\%) of ResNet-50 and ResNet-36 + attention pooling after finetuning from pretrained weights, found by \grover{} on limited and full labeled sets. The gain ($\Delta$) indicates how much improvement is made from using attention pooling in place of the last convolution block.}
  \label{tab:modela}
  \centering\small
  \begin{tabular}{*4c}
    \toprule
     Method & ResNet-50 & ResNet-36 + attention pooling & $\Delta$ \\
    \midrule
    
    CIFAR-10 8\% & 80.3 $\pm$ 0.3 & 81.3 $\pm$ 0.1 & {\bf +1.0}\\
    ImageNet 10\% & 61.8 $\pm$ 0.2 & 62.1 $\pm$ 0.2 & {\bf +0.3}\\
    \midrule
    CIFAR-10 100\% & 95.7 $\pm$ 0.1 & 95.4 $\pm$ 0.2 & -0.3\\
    ImageNet 100\% & 77.0 $\pm$ 0.1 & 77.5 $\pm$ 0.1 & {\bf +0.5}\\
    \bottomrule
  \end{tabular}
\end{table}

With pretraining on unlabeled data, ResNet-36 + attention pooling outperforms ResNet-50 on both datasets with limited data. On the full training set, this hybrid convolution-attention architecture gives 0.5\% gain on ImageNet $224\times224$. These show great promise for this hybrid architecture which we plan to further explore in future work.

\section{Related Work}
\label{sec:related}

\paragraph{Unsupervised representation learning for text.} Much of the success in unsupervised representation learning is in NLP. First, using language models to learn embeddings for words is commonplace in many NLP applications~\citep{mikolov2013distributed,pennington2014glove}. Building on this success, similar methods are then proposed for sentence and paragraph representations~\citep{le2014distributed,kiros2015skip}. Recent successful methods however focus on the use of language models or ``masked'' language models as pretraining objectives~\citep{dai2015semi,ramachandran2016unsupervised,peters2018deep,howard2018universal,devlin2018bert}. A general principle to all of these successful methods is the idea of context prediction: given some adjacent data and their locations, predict the missing words. 

\paragraph{Unsupervised representation learning for images and audio.} Recent successful methods in unsupervised representation learning for images can be divided into four categories: 1) predicting rotation angle from an original image (e.g.,~\citet{gidaris2018unsupervised}), 2) predicting if a perturbed image belongs to the same category with an unperturbed image (Exemplar) (e.g.,~\citet{dosovitskiy2016discriminative}), 3) predicting relative locations of patches (e.g.,~\citet{doersch2015unsupervised}),  solving Jigsaw puzzles (e.g.,~\citet{noroozi2016unsupervised}) and 4) impainting (e.g.,~\citet{huang2014image,pathak2016context,Iizuka:2017:GLC:3072959.3073659}). Their success, however, is limited to small datasets or small settings, some resort to expensive jointing training to surpass their purely supervised counterpart. On the challenging benchmark ImageNet, our method is the first to report gain with and without additional unlabeled data as shown in Table~\ref{tab:result}.

\grover{}
is also closely related to denoising autoencoders~\citep{vincent2010stacked}, where various kinds of noise are applied to the input and the model is required to reconstruct the clean input. The main difference between our method and denoising autoencoders is how the reconstruction step is done: our method focuses only on the missing patches, and tries to select the right patch among other distracting patches. Our method is also making use of Contrastive Predictive Coding~\citep{oord2018representation}, where negative sampling was also used to classify continuous objects in a sequence. 
Concurrent to our work, wav2vec~\citep{schneider2019wav2vec} also employs Contrastive Predictive Coding to learn representation of audio data during pretraining and achieve improvement on downstream tasks.

\paragraph{Semi-supervised learning.} Semi-supervised learning is another branch of representation learning methods that take advantage of the existence of labeled data. Unlike pure unsupervised representation learning, semi-supervised learning does not need a separate fine-tuning stage to improve  accuracy, which is more common in unsupervised representation learning. Successful recent semi-supervised learning methods for deep learning are based on consistency training~\citep{miyato2018virtual,sajjadi2016regularization,laine2016temporal,ver19,xie19}.

\section{Conclusion}
We introduce \grover{}, a self-supervised pretraining technique that generalizes
the concept of masked language modeling to continuous data, such as images. Given a masked-out position of a square patch in the input image, our method learns to select the target masked patches from negative samples obtained from the same image. This classification objective therefore sidesteps the need for predicting the exact pixel values of the target patches.
Experiments show that \grover{}
achieves significant gains when  labeled set is small compared to the unlabeled set. Besides the gain in mean accuracy across different runs, the standard deviation of results is also reduced thanks to a better initialization from our pretraining method. Our analysis demonstrates the revived potential of unsupervised pretraining over supervised learning and that a hybrid convolution-attention architecture shows promise.
%We also hope to address the caveats of our method in subsequent works.
\bibliography{main}

\begin{thebibliography}{40}
\expandafter\ifx\csname natexlab\endcsname\relax\def\natexlab#1{#1}\fi

\bibitem[{Bengio et~al.(2007)Bengio, Lamblin, Popovici, and
  Larochelle}]{bengio2007greedy}
Yoshua Bengio, Pascal Lamblin, Dan Popovici, and Hugo Larochelle. 2007.
\newblock Greedy layer-wise training of deep networks.
\newblock In \emph{Advances in Neural Information Processing Systems}, pages
  153--160.

\bibitem[{Cubuk et~al.(2018)Cubuk, Zoph, Mane, Vasudevan, and
  Le}]{cubuk2018autoaugment}
Ekin~D Cubuk, Barret Zoph, Dandelion Mane, Vijay Vasudevan, and Quoc~V Le.
  2018.
\newblock Autoaugment: Learning augmentation policies from data.
\newblock \emph{Proceedings of the IEEE conference on computer vision and
  pattern recognition}.

\bibitem[{Dai and Le(2015)}]{dai2015semi}
Andrew~M Dai and Quoc~V Le. 2015.
\newblock Semi-supervised sequence learning.
\newblock In \emph{Advances in Neural Information Processing Systems}, pages
  3079--3087.

\bibitem[{Devlin et~al.(2019)Devlin, Chang, Lee, and
  Toutanova}]{devlin2018bert}
Jacob Devlin, Ming-Wei Chang, Kenton Lee, and Kristina Toutanova. 2019.
\newblock {BERT}: Pre-training of deep bidirectional transformers for language
  understanding.
\newblock In \emph{Annual Conference of the North American Chapter of the
  Association for Computational Linguistics}.

\bibitem[{Doersch et~al.(2015)Doersch, Gupta, and
  Efros}]{doersch2015unsupervised}
Carl Doersch, Abhinav Gupta, and Alexei~A Efros. 2015.
\newblock Unsupervised visual representation learning by context prediction.
\newblock In \emph{Proceedings of the IEEE International Conference on Computer
  Vision}, pages 1422--1430.

\bibitem[{Dosovitskiy et~al.(2016)Dosovitskiy, Fischer, Springenberg,
  Riedmiller, and Brox}]{dosovitskiy2016discriminative}
Alexey Dosovitskiy, Philipp Fischer, Jost~Tobias Springenberg, Martin
  Riedmiller, and Thomas Brox. 2016.
\newblock Discriminative unsupervised feature learning with exemplar
  convolutional neural networks.
\newblock \emph{IEEE transactions on pattern analysis and machine
  intelligence}, 38(9):1734--1747.

\bibitem[{Gidaris et~al.(2018)Gidaris, Singh, and
  Komodakis}]{gidaris2018unsupervised}
Spyros Gidaris, Praveer Singh, and Nikos Komodakis. 2018.
\newblock Unsupervised representation learning by predicting image rotations.
\newblock In \emph{International Conference on Learning Representations}.

\bibitem[{Hannun et~al.(2014)Hannun, Case, Casper, Catanzaro, Diamos, Elsen,
  Prenger, Satheesh, Sengupta, Coates et~al.}]{hannun2014deep}
Awni Hannun, Carl Case, Jared Casper, Bryan Catanzaro, Greg Diamos, Erich
  Elsen, Ryan Prenger, Sanjeev Satheesh, Shubho Sengupta, Adam Coates, et~al.
  2014.
\newblock Deep speech: Scaling up end-to-end speech recognition.
\newblock \emph{arXiv preprint arXiv:1412.5567}.

\bibitem[{He et~al.(2016{\natexlab{a}})He, Zhang, Ren, and Sun}]{he2016deep}
Kaiming He, Xiangyu Zhang, Shaoqing Ren, and Jian Sun. 2016{\natexlab{a}}.
\newblock Deep residual learning for image recognition.
\newblock In \emph{Proceedings of the IEEE conference on computer vision and
  pattern recognition}, pages 770--778.

\bibitem[{He et~al.(2016{\natexlab{b}})He, Zhang, Ren, and
  Sun}]{he2016identity}
Kaiming He, Xiangyu Zhang, Shaoqing Ren, and Jian Sun. 2016{\natexlab{b}}.
\newblock Identity mappings in deep residual networks.
\newblock In \emph{European conference on computer vision}, pages 630--645.
  Springer.

\bibitem[{H{\'{e}}naff et~al.(2019)H{\'{e}}naff, Razavi, Doersch, Eslami, and
  van~den Oord}]{cpc19}
Olivier~J. H{\'{e}}naff, Ali Razavi, Carl Doersch, S.~M.~Ali Eslami, and
  A{\"{a}}ron van~den Oord. 2019.
\newblock Data-efficient image recognition with contrastive predictive coding.
\newblock \emph{CoRR}, abs/1905.09272.

\bibitem[{Hinton et~al.(2006)Hinton, Osindero, and Teh}]{hinton2006fast}
Geoffrey~E Hinton, Simon Osindero, and Yee-Whye Teh. 2006.
\newblock A fast learning algorithm for deep belief nets.
\newblock \emph{Neural Computation}, 18(7):1527--1554.

\bibitem[{Hjelm et~al.(2018)Hjelm, Fedorov, Lavoie-Marchildon, Grewal,
  Trischler, and Bengio}]{hjelm2018learning}
R~Devon Hjelm, Alex Fedorov, Samuel Lavoie-Marchildon, Karan Grewal, Adam
  Trischler, and Yoshua Bengio. 2018.
\newblock Learning deep representations by mutual information estimation and
  maximization.
\newblock \emph{arXiv preprint arXiv:1808.06670}.

\bibitem[{Howard and Ruder(2018)}]{howard2018universal}
Jeremy Howard and Sebastian Ruder. 2018.
\newblock Universal language model fine-tuning for text classification.
\newblock In \emph{Annual Conference of the North American Chapter of the
  Association for Computational Linguistics}.

\bibitem[{Huang et~al.(2017)Huang, Liu, Van Der~Maaten, and
  Weinberger}]{huang2017densely}
Gao Huang, Zhuang Liu, Laurens Van Der~Maaten, and Kilian~Q Weinberger. 2017.
\newblock Densely connected convolutional networks.
\newblock In \emph{Proceedings of the IEEE conference on computer vision and
  pattern recognition}, pages 4700--4708.

\bibitem[{Huang et~al.(2014)Huang, Kang, Ahuja, and Kopf}]{huang2014image}
Jia-Bin Huang, Sing~Bing Kang, Narendra Ahuja, and Johannes Kopf. 2014.
\newblock Image completion using planar structure guidance.
\newblock \emph{ACM Transactions on graphics (TOG)}, 33(4):129.

\bibitem[{Iizuka et~al.(2017)Iizuka, Simo-Serra, and
  Ishikawa}]{Iizuka:2017:GLC:3072959.3073659}
Satoshi Iizuka, Edgar Simo-Serra, and Hiroshi Ishikawa. 2017.
\newblock Globally and locally consistent image completion.
\newblock \emph{ACM Trans. Graph.}, 36(4):107:1--107:14.

\bibitem[{Kiros et~al.(2015)Kiros, Zhu, Salakhutdinov, Zemel, Urtasun,
  Torralba, and Fidler}]{kiros2015skip}
Ryan Kiros, Yukun Zhu, Ruslan~R Salakhutdinov, Richard Zemel, Raquel Urtasun,
  Antonio Torralba, and Sanja Fidler. 2015.
\newblock Skip-thought vectors.
\newblock In \emph{Advances in Neural Information Processing Systems}, pages
  3294--3302.

\bibitem[{Kolesnikov et~al.(2019)Kolesnikov, Zhai, and
  Beyer}]{kolesnikov2019revisiting}
Alexander Kolesnikov, Xiaohua Zhai, and Lucas Beyer. 2019.
\newblock Revisiting self-supervised visual representation learning.
\newblock \emph{arXiv preprint arXiv:1901.09005}.

\bibitem[{Laine and Aila(2016)}]{laine2016temporal}
Samuli Laine and Timo Aila. 2016.
\newblock Temporal ensembling for semi-supervised learning.
\newblock In \emph{International Conference on Learning Representations}.

\bibitem[{Le and Mikolov(2014)}]{le2014distributed}
Quoc Le and Tomas Mikolov. 2014.
\newblock Distributed representations of sentences and documents.
\newblock In \emph{International Conference on Machine Learning}, pages
  1188--1196.

\bibitem[{Mikolov et~al.(2013)Mikolov, Sutskever, Chen, Corrado, and
  Dean}]{mikolov2013distributed}
Tomas Mikolov, Ilya Sutskever, Kai Chen, Greg~S Corrado, and Jeff Dean. 2013.
\newblock Distributed representations of words and phrases and their
  compositionality.
\newblock In \emph{Advances in Neural Information Processing Systems}, pages
  3111--3119.

\bibitem[{Miyato et~al.(2018)Miyato, Maeda, Ishii, and
  Koyama}]{miyato2018virtual}
Takeru Miyato, Shin-ichi Maeda, Shin Ishii, and Masanori Koyama. 2018.
\newblock Virtual adversarial training: a regularization method for supervised
  and semi-supervised learning.
\newblock \emph{IEEE transactions on pattern analysis and machine
  intelligence}.

\bibitem[{Noroozi and Favaro(2016)}]{noroozi2016unsupervised}
Mehdi Noroozi and Paolo Favaro. 2016.
\newblock Unsupervised learning of visual representations by solving jigsaw
  puzzles.
\newblock In \emph{European Conference on Computer Vision}, pages 69--84.
  Springer.

\bibitem[{Oliver et~al.(2018)Oliver, Odena, Raffel, Cubuk, and
  Goodfellow}]{oliver2018realistic}
Avital Oliver, Augustus Odena, Colin~A Raffel, Ekin~Dogus Cubuk, and Ian
  Goodfellow. 2018.
\newblock Realistic evaluation of deep semi-supervised learning algorithms.
\newblock In \emph{Advances in Neural Information Processing Systems}, pages
  3235--3246.

\bibitem[{Oord et~al.(2018)Oord, Li, and Vinyals}]{oord2018representation}
Aaron van~den Oord, Yazhe Li, and Oriol Vinyals. 2018.
\newblock Representation learning with contrastive predictive coding.
\newblock \emph{arXiv preprint arXiv:1807.03748}.

\bibitem[{Pathak et~al.(2016)Pathak, Krahenbuhl, Donahue, Darrell, and
  Efros}]{pathak2016context}
Deepak Pathak, Philipp Krahenbuhl, Jeff Donahue, Trevor Darrell, and Alexei~A
  Efros. 2016.
\newblock Context encoders: Feature learning by inpainting.
\newblock In \emph{Proceedings of the IEEE conference on computer vision and
  pattern recognition}, pages 2536--2544.

\bibitem[{Pennington et~al.(2014)Pennington, Socher, and
  Manning}]{pennington2014glove}
Jeffrey Pennington, Richard Socher, and Christopher Manning. 2014.
\newblock Glove: Global vectors for word representation.
\newblock In \emph{Proceedings of the 2014 conference on empirical methods in
  natural language processing (EMNLP)}, pages 1532--1543.

\bibitem[{Peters et~al.(2018)Peters, Neumann, Iyyer, Gardner, Clark, Lee, and
  Zettlemoyer}]{peters2018deep}
Matthew~E Peters, Mark Neumann, Mohit Iyyer, Matt Gardner, Christopher Clark,
  Kenton Lee, and Luke Zettlemoyer. 2018.
\newblock Deep contextualized word representations.
\newblock In \emph{Annual Conference of the North American Chapter of the
  Association for Computational Linguistics}.

\bibitem[{Raina et~al.(2007)Raina, Battle, Lee, Packer, and Ng}]{raina2007self}
Rajat Raina, Alexis Battle, Honglak Lee, Benjamin Packer, and Andrew~Y Ng.
  2007.
\newblock Self-taught learning: transfer learning from unlabeled data.
\newblock In \emph{Proceedings of the 24th International Conference on Machine
  Learning}, pages 759--766. ACM.

\bibitem[{Ramachandran et~al.(2017)Ramachandran, Liu, and
  Le}]{ramachandran2016unsupervised}
Prajit Ramachandran, Peter~J Liu, and Quoc~V Le. 2017.
\newblock Unsupervised pretraining for sequence to sequence learning.
\newblock In \emph{Proceedings of the 2017 Conference on Empirical Methods in
  Natural Language Processing}.

\bibitem[{Sajjadi et~al.(2016)Sajjadi, Javanmardi, and
  Tasdizen}]{sajjadi2016regularization}
Mehdi Sajjadi, Mehran Javanmardi, and Tolga Tasdizen. 2016.
\newblock Regularization with stochastic transformations and perturbations for
  deep semi-supervised learning.
\newblock In \emph{Advances in Neural Information Processing Systems}, pages
  1163--1171.

\bibitem[{Schneider et~al.(2019)Schneider, Baevski, Collobert, and
  Auli}]{schneider2019wav2vec}
Steffen Schneider, Alexei Baevski, Ronan Collobert, and Michael Auli. 2019.
\newblock wav2vec: Unsupervised pre-training for speech recognition.
\newblock \emph{arXiv preprint arXiv:1904.05862}.

\bibitem[{Vaswani et~al.(2017)Vaswani, Shazeer, Parmar, Uszkoreit, Jones,
  Gomez, Kaiser, and Polosukhin}]{vaswani2017attention}
Ashish Vaswani, Noam Shazeer, Niki Parmar, Jakob Uszkoreit, Llion Jones,
  Aidan~N Gomez, {\L}ukasz Kaiser, and Illia Polosukhin. 2017.
\newblock Attention is all you need.
\newblock In \emph{Advances in Neural Information Processing Systems}, pages
  5998--6008.

\bibitem[{Verma et~al.(2019)Verma, Lamb, Kannala, Bengio, and
  Lopez-Paz}]{ver19}
Vikas Verma, Alex Lamb, Juho Kannala, Yoshua Bengio, and David Lopez-Paz. 2019.
\newblock Interpolation consistency training for semi-supervised learning.
\newblock \emph{arXiv preprint arXiv:1903.03825}.

\bibitem[{Vincent et~al.(2010)Vincent, Larochelle, Lajoie, Bengio, and
  Manzagol}]{vincent2010stacked}
Pascal Vincent, Hugo Larochelle, Isabelle Lajoie, Yoshua Bengio, and
  Pierre-Antoine Manzagol. 2010.
\newblock Stacked denoising autoencoders: Learning useful representations in a
  deep network with a local denoising criterion.
\newblock \emph{Journal of machine learning research}, 11(Dec):3371--3408.

\bibitem[{Wu et~al.(2016)Wu, Schuster, Chen, Le, Norouzi, Macherey, Krikun,
  Cao, Gao, Macherey et~al.}]{wu2016google}
Yonghui Wu, Mike Schuster, Zhifeng Chen, Quoc~V Le, Mohammad Norouzi, Wolfgang
  Macherey, Maxim Krikun, Yuan Cao, Qin Gao, Klaus Macherey, et~al. 2016.
\newblock Google's neural machine translation system: Bridging the gap between
  human and machine translation.
\newblock \emph{arXiv preprint arXiv:1609.08144}.

\bibitem[{Xie et~al.(2019)Xie, Dai, Hovy, Luong, and Le}]{xie19}
Qizhe Xie, Zihang Dai, Eduard Hovy, Minh-Thang Luong, and Quoc~V. Le. 2019.
\newblock Unsupervised data augmentation.
\newblock \emph{arXiv preprint arXiv:1904.12848}.

\bibitem[{Zagoruyko and Komodakis(2016)}]{zagoruyko2016wide}
Sergey Zagoruyko and Nikos Komodakis. 2016.
\newblock Wide residual networks.
\newblock In \emph{The British Machine Vision Conference}.

\bibitem[{Zhai et~al.(2019)Zhai, Oliver, Kolesnikov, and Beyer}]{zhai2019s}
Xiaohua Zhai, Avital Oliver, Alexander Kolesnikov, and Lucas Beyer. 2019.
\newblock $s^4l$: Self-supervised semi-supervised learning.
\newblock \emph{arXiv preprint arXiv:1905.03670}.

\end{thebibliography}

\end{document}